\documentclass[sigconf]{acmart}

\renewcommand\footnotetextcopyrightpermission[1]{}
\usepackage{hyperref}

\AtBeginDocument{%
  }

\setcopyright{none}
\begin{document}

\title{MNN-LLM: A Generic Inference Engine for Fast Large Language Model Deployment on Mobile Devices}


\author{Zhaode Wang}
\affiliation{%
  \institution{Alibaba Group}
  \city{Beijing}
  \country{China}
}
\email{zhaode.wzd@taobao.com}

\author{Jingbang Yang}
\affiliation{%
  \institution{Alibaba Group}
  \city{Hangzhou}
  \country{China}
}
\email{jingbang.yjb@taobao.com}

\author{Xinyu Qian}
\affiliation{%
  \institution{Alibaba Group}
  \city{Hangzhou}
  \country{China}
}
\email{qianxinyu.qxy@taobao.com}

\author{Shiwen Xing}
\affiliation{%
  \institution{Alibaba Group}
  \city{Hangzhou}
  \country{China}
}
\email{tianbu.xsw@taobao.com}

\author{Xiaotang Jiang}
\affiliation{%
  \institution{Alibaba Group}
  \city{Hangzhou}
  \country{China}
}
\email{xiaotang.jxt@taobao.com}

\author{Chengfei Lv}
\authornote{Chengfei Lv is the corresponding author.}
\affiliation{%
  \institution{Alibaba Group}
  \city{Hangzhou}
  \country{China}
}
\email{chengfei.lcf@taobao.com}

\author{Shengyu Zhang}
\affiliation{%
  \institution{Zhejiang University}
  \city{Hangzhou}
  \country{China}
}
\email{sy_zhang@zju.edu.cn}
\thanks{%
  \parbox{\textwidth}{\footnotesize
    \textit MM Asia 2024
  }
}

% \renewcommand{\shortauthors}{Zhaode Wang et al.}
% \renewcommand{\shortauthors}{Anonymous Author, et al.}

\begin{abstract}
  Large language models (LLMs) have demonstrated exceptional performance across a variety of tasks. However, their substantial scale leads to significant computational resource consumption during inference, resulting in high costs. Consequently, edge device inference presents a promising solution. The primary challenges of edge inference include memory usage and inference speed. This paper introduces MNN-LLM, a framework specifically designed to accelerate the deployment of large language models on mobile devices. MNN-LLM addresses the runtime characteristics of LLMs through model quantization and DRAM-Flash hybrid storage, effectively reducing memory usage. It rearranges weights and inputs based on mobile CPU instruction sets and GPU characteristics while employing strategies such as multicore load balancing, mixed-precision floating-point operations, and geometric computations to enhance performance. Notably, MNN-LLM achieves up to a 8.6x speed increase compared to current mainstream LLM-specific frameworks.
\end{abstract}

\maketitle
\pagestyle{plain}

\section{Introduction}
In recent years, large language models (LLMs) have rapidly evolved, becoming one of the most revolutionary technologies in the field of natural language processing. Prominent models like ChatGPT-4o \cite{OpenAI23} and QwenMax \cite{qwen}, due to their substantial parameter scales, are typically deployed in the cloud using GPU inference. However, the large parameter size and computational demands lead to high user costs, and utilizing cloud services may involve handling sensitive user data, raising privacy and security concerns. In response to these issues, the trend of deploying LLMs on mobile devices is gaining momentum \cite{aiondevice}. While mobile devices are widely used, their memory and computational limitations make it challenging to run large-scale LLMs directly.

To address this challenge, many open-source pre-trained LLMs have released smaller-scale models tailored for edge environments. These models share the same architecture as their cloud counterparts but have fewer parameters, such as Qwen2-1.5B \cite{qwen2}. According to the Scaling Law \cite{scaling20}, smaller models exhibit limited capabilities and often can only perform simple or specific tasks. However, with the advent of models like phi-1 \cite{phi123}, it has been shown that the quality of training data can significantly enhance the capabilities of smaller parameter models. Additionally, reinforcement learning algorithms in models like ChatGPT-o1 \cite{OpenAI23} can further improve LLM performance without increasing the parameter count. As a result, the gap between smaller and larger models is gradually narrowing; for instance, the 3B parameter Qwen2.5-3B \cite{qwen2.5} achieved a score of 64 on the MMLU benchmark, surpassing many 30B parameter predecessors. This rapid enhancement in LLM capabilities offers greater feasibility for deploying LLMs on mobile devices.

Despite the reduction in parameter scale, the computational demands of LLMs remain substantial compared to traditional computer vision models used for edge inference. Running LLMs smoothly on edge devices poses significant challenges due to hardware, memory, and computational constraints. To address this issue, several frameworks for deploying LLMs on mobile devices have emerged. Some frameworks, such as PowerInfer-2 \cite{powerinfer2} and LLM in Flash \cite{llminflash}, require modifications to the model for compatibility. Others, like llama.cpp \cite{llama.cpp}, MLC-LLM \cite{mlc-llm}, and fastllm \cite{fastllm}, can be used directly with a focus on LLMs.

This paper introduces MNN-LLM, a generic mobile inference framework that supports LLM inference as well as the deployment of various deep learning models. MNN-LLM is developed based on MNN \cite{MNN20}, a general framework designed for executing deep learning model inference on mobile devices. It addresses the challenges posed by large-scale LLMs through targeted optimizations in model export, quantization, and computational graph optimization. Furthermore, MNN-LLM analyzes the inference process and employs various forms of combined quantization, utilizing DRAM-Flash hybrid storage to reduce runtime memory usage. By optimizing high-computation operators and rearranging data according to the characteristics of different hardware, MNN-LLM ensures optimal utilization of edge computing resources.

\section{Background and Motivation}

\subsection{LLM Model and Inference}

Currently, mainstream LLMs primarily adopt a Decoder-Only architecture, with the main parameters located in the Embedding and Linear operators. During inference, the operators that consume the most time are Linear and Attention. 

The inference process can be divided into two phases: prefill and decode. The prefill phase refers to the computation of the input text, processing the user's input text sequence to generate the first token. The decode phase involves generating text, where each decode operation produces one token until a termination token is generated. These two phases exhibit different computational characteristics; specifically, the prefill phase tends to be computation-bound, while the decode phase is memory-bound due to the typical computational throughput and memory bandwidth of edge devices. 

In the inference phase of the Attention mechanism, there are three inputs: query, key, and value. During the decode phase, only the query generated from the current input token is needed; however, all previously computed keys and values are required. To reduce computational load and avoid redundant calculations, a key-value (KV) cache is typically employed to store the keys and values from prior computations.

\subsection{Mobile Devices Analysis}

% Mobile devices typically have limited memory capacity, and the large parameter sizes of LLMs require substantial memory usage. Additionally, as the context length increases during generation, the KV cache also consumes significant memory. This can lead to memory shortages on mobile devices, potentially resulting in processes being terminated. While the read speed of Flash memory is considerably slower than that of DRAM, its capacity is often orders of magnitude greater, making it essential to consider the use of Flash during LLM inference.

% Mobile devices feature computing hardware such as CPUs and GPUs. Mobile CPUs typically have multiple cores (ranging from 4 to 8) and often follow a big.LITTLE \cite{biglittle} architecture, making concurrent optimization essential during CPU development. Additionally, differences in instruction sets between various CPUs can lead to distinct performance optimizations; thus, specific optimizations for different instruction sets are necessary to achieve optimal performance. General-purpose computing on mobile GPUs typically utilizes Vulkan and OpenCL standards. Developers can parallelize computational tasks using the APIs provided by these standards and segment parallel tasks according to the specified abstract execution model. The hardware drivers automatically handle the specific instruction dispatch and task scheduling, facilitating code portability across different platforms.

Mobile devices often have limited memory, and the large parameter sizes of LLMs can lead to significant memory usage, especially as context length increases, resulting in memory shortages that may terminate processes. Although Flash memory has much slower read speeds than DRAM, its higher capacity makes it crucial for LLM inference.

Mobile devices utilize CPUs and GPUs, typically featuring multiple cores and often following a big.LITTLE \cite{biglittle} architecture, necessitating concurrent optimization during CPU development. Variations in instruction sets across CPUs require tailored optimizations for optimal performance. General-purpose computing on mobile GPUs typically employs Vulkan and OpenCL standards, allowing developers to parallelize tasks using these APIs. Hardware drivers manage instruction dispatch and task scheduling, enhancing code portability across platforms.

\section{MNN-LLM Overview}

MNN-LLM, built on the deep learning framework MNN, leverages MNN's extensive operator set and model supported versatility to enhance flexibility and adaptability. Unlike LLM-specific inference engines, MNN-LLM supports a wider range of models, boosting usability and developer friendliness in edge computing contexts.

MNN provides robust support for computer vision (CV) models like MobileNet \cite{mobilenet} and YOLO \cite{yolo}, which are typically exported to ONNX \cite{onnx} before conversion to MNN format. While MNN-LLM uses these export and conversion processes, the large parameter sizes of LLMs can result in high memory usage and longer conversion times. To mitigate this, optimizations have been introduced: Linear operators are replaced with custom operators during graph export, allowing ONNX export to focus on the computation graph without parameters. After model export, conversion, and optimization, parameters can be handled separately, streamlining the process and leveraging MNN's model format. Additionally, during model conversion, optimizations such as RMSNorm \cite{rmsnorm} fusion and Attention fusion are applied. The computation graph also supports the runtime loading of LoRA \cite{lora} weights, enabling seamless integration of LoRA models without requiring external implementations in the inference framework.

MNN-LLM provides robust quantization capabilities, supporting both integer (int) and floating-point (fp) quantization during the model conversion phase, as well as quantization of activation values and KV cache during runtime. Additionally, it supports other quantization algorithms, such as GPTQ\cite{gptq}, and allows for the import of quantized weights.

MNN-LLM performs extensive optimizations for memory and computation at runtime. To address the significant memory usage of LLMs, it employs methods such as DRAM-Flash hybrid storagea and combined quantization. For the high computational load, strategies like data reorder tailored for hardware, multicore load balancing, mixed-precision floating-point operations, and geometry computations are utilized. Additionally, specific optimizations are implemented for multi-LoRA scenarios.

\section{Memory Optimization}

\subsection{DRAM-Flash Hybrid Storage}

\begin{figure}[ht]
  \centering
  \includegraphics[width=\linewidth]{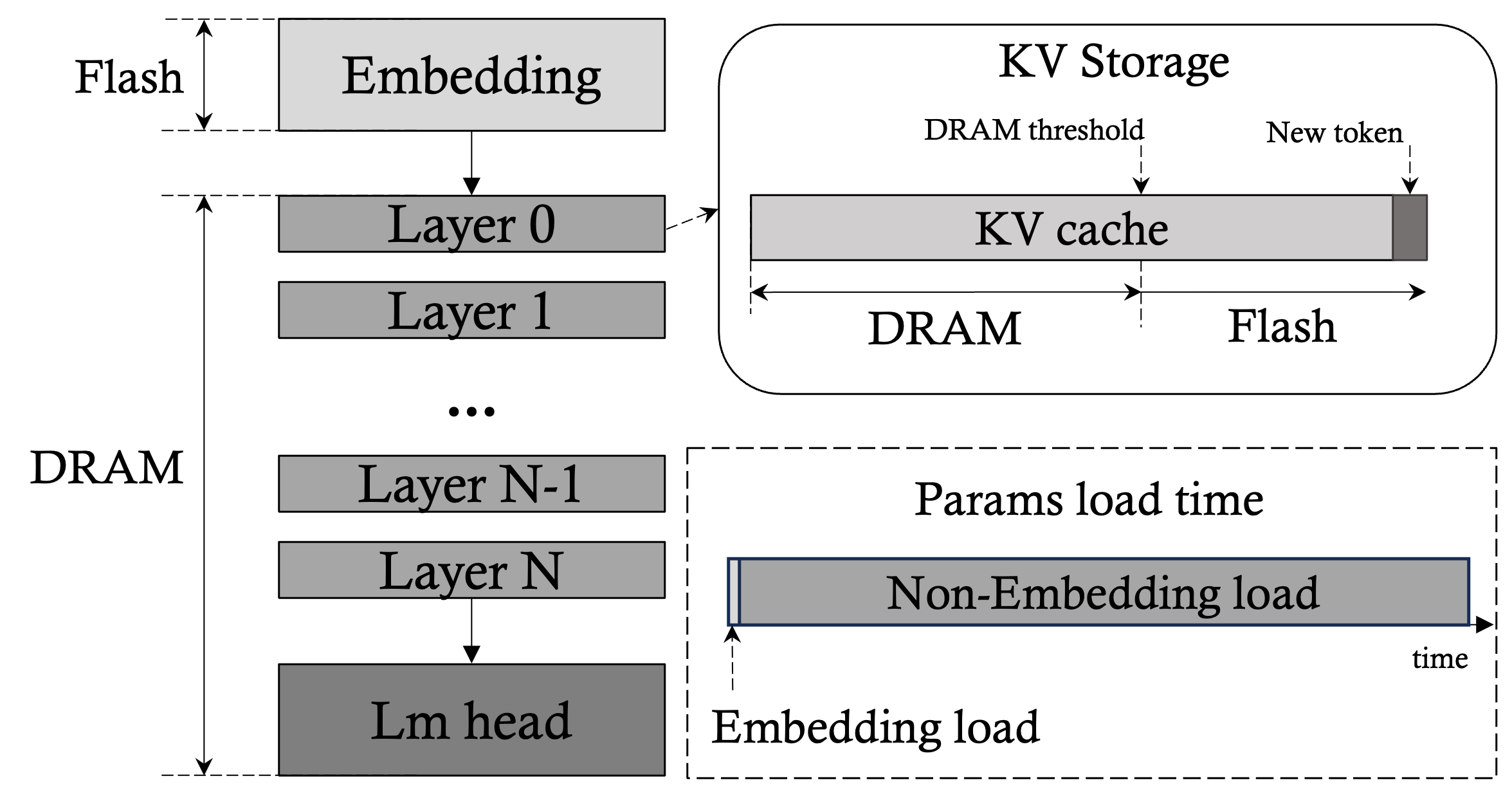}
  \caption{DRAM-Flash Hybrid Storage for LLM model parameters and KV cache.}
  \label{fig:hybrid}
\end{figure}

The primary bottleneck for deploying large LLM models on mobile devices lies in the limitations of DRAM. MNN-LLM employs a DRAM-Flash hybrid storage strategy to mitigate memory usage, ensuring minimal memory occupancy while maintaining the usability of LLM inference under constrained memory conditions. Although Flash storage has a larger capacity than DRAM, its read speeds are significantly slower; for instance, LPDDR5X achieves approximately 58 GB/s, while UFS 4.0 ranges from about 450 MB/s to 3 GB/s \cite{powerinfer2}. This means that DRAM can be 19 to 130 times faster than Flash. While hybrid storage can reduce memory demands and enhance usability, it may compromise inference performance. As shown in Figure~\ref{fig:hybrid}, MNN-LLM's hybrid storage strategy is tailored to the operational characteristics of the model: for parameter storage, it assesses utilization rates and allocates low-utilization parameters to Flash to minimize speed impact. For the KV data, prefetching techniques are employed to reduce the latency of Flash reads, thereby mitigating their effect on performance.

\begin{table}[ht]
    \caption{Qwen2 7B Model Params}
    \label{tab:params}
    \centering
    \begin{tabular}{ccl}
        \begin{tabular}{@{}cc@{}}
            \toprule
            Configuration&Size\\
            \midrule
            vocabulary size & 151646\\
            hidden size & 3584\\
            intermediate size & 18944\\
            layers & 28\\
            \bottomrule
        \end{tabular} & \hspace{1cm}
        \begin{tabular}{@{}cc@{}}
            \toprule
            Params&Size\\
            \midrule
            Embedding & 1.09 B\\
            Layers & 4.89 B\\
            Lm head & 1.09 B\\
            Total & 7.07 B\\
            \bottomrule
        \end{tabular} \\
    \end{tabular}
\end{table}

The large parameter scale of LLM models is a primary reason for their high memory consumption. Structurally, the parameters can be divided into three categories: \textit{Embedding}, \textit{Layer}, and \textit{Lm head}. The size of the \textit{Embedding} and \textit{Lm head} parameters is generally calculated as $\textit{vocabulary size} \times \textit{hidden size}$, and since the \textit{vocabulary size} is usually large, the \textit{Embedding} parameters do not participate in calculations like other parameters do. \textit{Layer} refer to the parameters in each Decoder Layer, including the Attention and MLP Linear layers, typically sized at $\textit{hidden size} \times \textit{hidden size}$ or $\textit{intermediate size} \times \textit{hidden size}$ with consistent parameter scales across layers. As shown in Table~\ref{tab:params}, in the Qwen2 7B \cite{qwen2}  model, the non-computational \textit{Embedding} parameters account for about 15\% of the total parameters.

In the decode phase, each input consists of the previously generated token. Leading to a computational process that necessitates loading 1/vocabulary size of the \textit{Embedding} parameters, along with full \textit{Layer} and \textit{Lm head} parameters. Thus, \textit{Layer} and \textit{Lm head} parameters should be prioritized for DRAM storage, while the \textit{Embedding} parameters can be stored in Flash. Taking Qwen2 7B as an example, with Embedding data read in bfloat16 format, the decode phase only requires the Embedding value for one token, resulting in a data size of 7 KB for each decode. The UFS 4.0 read speed is approximately \(15 \mu s\) slower than LPDDR5X. In contrast, loading non-Embedding parameters from memory takes about 103 ms. In typical mobile devices, the compute characteristics during the decode phase are Memory Bound, making the memory access time roughly equivalent to the parameter access time. Therefore, storing \textit{Embedding} parameters in Flash adds only about 1.4 ‱ to the total inference time. Consequently, utilizing Flash for storing Embedding layers allows for a 15\% reduction in DRAM usage without significantly impacting inference performance. For example, Qwen-7B can reduce DRAM usage by approximately ~2.18 GB when using bfloat16 storage, greatly enhancing the feasibility of model inference on memory-constrained mobile devices.

% For the prefill phase, the number of tokens varies. Assuming there are $N$ tokens, the additional time incurred from using Flash instead of DRAM for the Embedding phase is about $15 \times N \mu s$. When $N$ is small, the time increase is comparable to that during the decode phase. As $N$ increases and the prefill becomes Compute Bound, total inference time rises based on computational throughput; however, the incremental time from Embedding remains negligible. 

\begin{figure}[h]
  \centering
  \includegraphics[width=\linewidth]{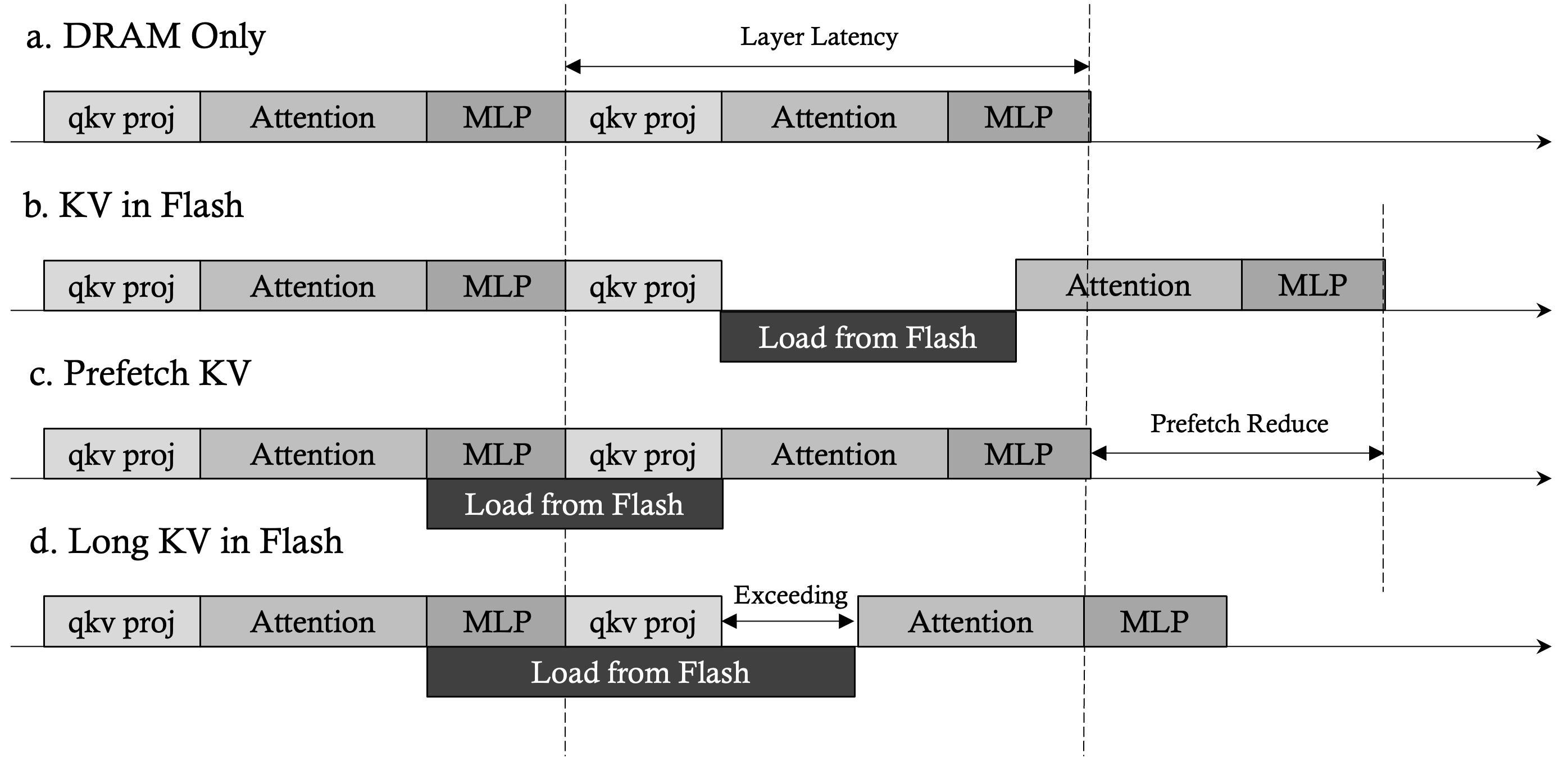}
  \caption{Comparison of KV loading times for DRAM, DRAM-Flash, Prefetching, and Exceeding.}
  \label{fig:kvflash}
\end{figure}

In scenarios with long input texts or extensive generation lengths, the continuous growth of the KV cache can lead to significant memory usage. MNN-LLM addresses this challenge by employing a hybrid storage strategy, utilizing Flash to hold part of the KV cache, thus ensuring LLM inference remains feasible under long-context conditions. Initially, all KV cache values are stored in DRAM, but as the context expands and the KV cache size increases, any portion exceeding a certain threshold is transferred to Flash. Since each computation produces only one set of new KV values, the total number of KV values for the Qwen2 7B model amounts to approximately 1 KB, minimizing storage overhead.

As the number of KV cache values stored in Flash rises, the time required to load them from Flash will gradually increase, which can slow down inference speed, as illustrated in Figure~\ref{fig:kvflash}b. To mitigate the impact of KV cache loading from Flash on inference time, we implement prefetching: during the MLP phase of the current layer and the qkv projection phase of the next layer, KV cache values are prefetched from Flash into memory. When the prefetching time is less than or equal to the computation time, the LLM inference speed remains unaffected.

For instance, in the Qwen2 7B model, the parameter size for a single layer's qkv and MLP is 178.83 MB, and the decode phase is Memory Bound. Given that LPDDR5X incurs about 3 ms of loading time for this data, we assume a loading speed of 1 GB/s for Flash due to its larger continuous memory blocks, allowing approximately 3 MB of KV values to be loaded within the computation time. Therefore, when the length of the KV cache stored in Flash is under 3072 K, the overhead from Flash loading is effectively masked by the computation time, as shown in Figure~\ref{fig:kvflash}c. However, once the length of the KV cache in Flash exceeds 3072 K, as depicted in Figure~\ref{fig:kvflash}d, prefetching cannot completely offset the Flash loading overhead; each additional 1 K of length adds approximately 1 ms of delay. It is important to note that DRAM also holds a substantial length of KV cache, meaning that only in scenarios with exceedingly long contexts will the inference speed of the LLM be impacted. Nevertheless, storing KV cache in Flash ensures that LLM inference remains viable even with long contexts.

\subsection{Combined Quantization}

The large parameter size of LLM models is the primary reason for their high memory consumption, and quantization can significantly reduce the parameter size, thereby lowering memory usage. However, quantization can affect the model's inference accuracy; generally, lower bit counts result in greater information loss and a larger impact on accuracy. There are various methods, data types, and bit counts for quantization, making it crucial to choose an appropriate method to balance memory usage, runtime performance, and model accuracy. 

For the parameters of the Embedding, Layer, and Lm head, MNN-LLM employs a combination quantization strategy to balance accuracy and computational overhead. The weights of the embedding layer account for approximately 15\% of the total model weight. Since only a small portion of these weights is utilized during each decoding step, they are stored in Flash memory, which does not occupy DRAM. This allows for the use of bfloat16 storage, ensuring computational accuracy. Non-embedding parameters, which include the weights of the layers and the LM head, must be fully loaded for each computation, making their size significantly impactful on inference performance. In particular, during the decoding phase, which is memory-bound, the inference time is directly proportional to the size of these parameters. Therefore, it is crucial to use low-bit quantization for these weights. Taking both precision and hardware computation instructions into account—where edge CPUs are particularly friendly towards int8 computation—these parameters are quantized using int4 or int8. During calculations, activation values are quantized to int8, enabling the use of W4A8 or W8A8 computation methods on CPUs to leverage int8 instructions. On GPUs, W4A16 or W8A16 methods are used to take advantage of floating-point capabilities. To maintain model accuracy, all these parameters employ asymmetric quantization. Asymmetric quantization as below: 
\begin{equation}
    w_{asy} = \textit{round}\left(\frac{w_{float} - w_{min}}{\frac{w_{max} - w_{min}}{clip_{max} - clip_{min}}}\right) + clip_{min}
\end{equation}
Additionally, because the LM head has a greater impact on model accuracy than the layers, it is prioritized for int8 quantization to enhance overall precision.

\begin{figure}[h]
  \centering
  \includegraphics[width=\linewidth]{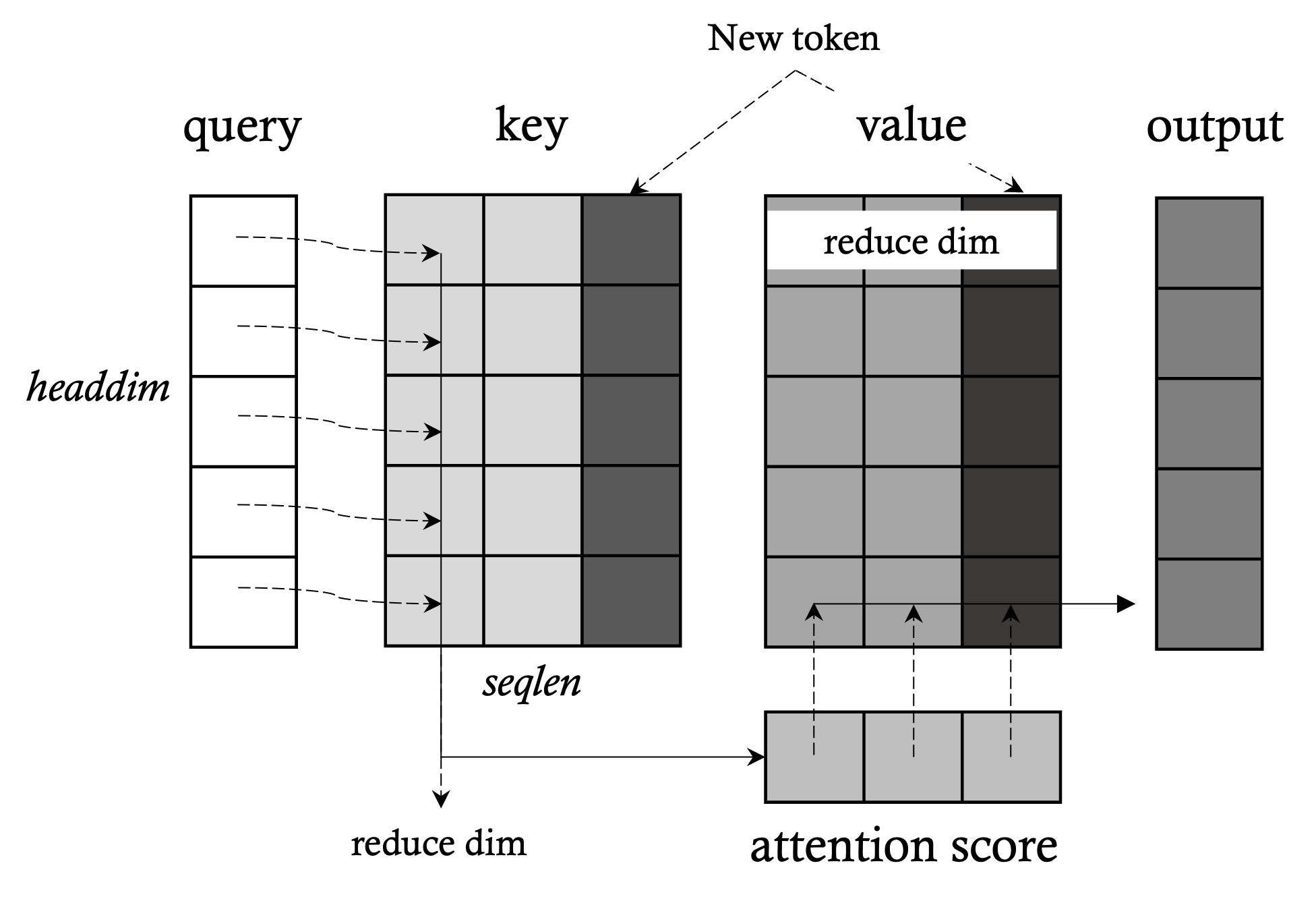}
  \caption{The reduction dimensions in the computation of Attention query, key, and value.}
  \label{fig:kvquant}
\end{figure}

When dealing with long contexts, the memory usage of the KV cache continues to grow, and quantization strategies can effectively reduce this memory consumption. MNN-LLM provides different quantization methods for keys and values based on their computational roles. During attention calculations, the shapes for query, key, and value are $\left[\textit{headnum}, \textit{seqlen}, \textit{headdim}\right]$. When performing matrix multiplication between key and query, the dimension being reduced is \textit{headdim}, which is a fixed value. Therefore, int4/int8 quantization can be applied to keys, allowing new key values to be quantized and stored directly. In contrast, during the matrix multiplication of \textit{attention score} with values, the dimension being reduced is \textit{seqlen}. Using int4/int8 quantization for values can affect the data distribution of existing values when new ones are added, necessitating updates to their quantization values and incurring additional overhead. To address this, MNN-LLM employs fp8 quantization for values, allowing new values to be quantized directly without impacting the existing ones.

\section{Compute Optimized}

\subsection{Hardware-Driven Data Reorder}

Analysis of the inference process in LLMs shows that the primary time-consuming operations are Linear and Attention, both of which fundamentally rely on matrix multiplication. Therefore, optimizing matrix multiplication for these two operators is crucial for improving LLM performance. Loop Tiling is a common optimization technique that enhances memory access locality, significantly impacting performance. The optimal tile size for Loop Tiling \cite{looptiling} greatly affects the final matrix multiplication performance and is influenced by the device's memory, cache, and computational hardware. Thus, it is essential to select the most suitable data reorganization and computation method based on hardware and data scale to achieve peak performance. MNN-LLM employs a Hardware-Driven data reorder strategy tailored to the computational characteristics of these two operator types to determine the tiling method and size, optimizing LLM inference performance.

\begin{table}
  \caption{Tile Sizes for Different CPU Architectures}
  \label{tab:tilesize}
  \begin{tabular}{cccc}
    \toprule
    Architecture & $e_p$ & $h_p$ & $l_p$ \\
    \midrule
    ARM i8sdot & 12 & 8 & 4 \\
    ARM i8mm & 10 & 8 & 8 \\
    X86 AVX2 & 4 & 8 & 4 \\
    X86 AVX512 & 4 & 64 & 4 \\
    \bottomrule
  \end{tabular}
\end{table}

The matrix multiplication for the Linear operator involves the activation values and weight values, where the activation values are computed during inference and the weights are determined when the model is loaded. In MNN-LLM, weights are generally quantized to int4 or int8. Assuming the activation value matrix size is $\left[e, l\right]$, and the weight size is $\left[h, l\right]$, the resulting size will be $\left[e, h\right]$. After data tiling on the mobile CPU, the input matrices are rearranged as: $\left[\frac{e}{e_p}, \frac{l}{l_p}, e_p, l_p\right]$ for the activation values and $\left[\frac{h}{h_p}, \frac{l}{l_p}, e_p, l_p\right]$ for the weights. This tiling allows for value reuse within the registers during kernel computations, enhancing memory locality and reducing memory access frequency. The memory access count is optimized from $2ehl + eh$ to $\frac{e}{e_p}\frac{h}{h_p}(le_p+lh_p+h_pe_p)$. By using memory access frequency as the optimization objective and hardware parameters as constraints, we can compute the values for $e_p, h_p, l_p$ under different hardware conditions. Let $R$ be the number of vector registers, and $instruction_{width}$ be the data size computed in a single instruction along the l-dimension, $e_p, h_p, l_p$ as given by the following formulas:
\begin{align}
\text{min} \quad & \frac{e}{e_p}\frac{h}{h_p}(le_p+lh_p+h_pe_p) \\
\text{s.t.} \quad & e_p + h_p + h_pe_p \le R \\
& l_p = instruction_{width}
\end{align}
Based on the above strategy, the block sizes calculated for various CPU instruction sets are shown in Table~\ref{tab:tilesize}. By employing a Hardware-Driven data rearrangement strategy tailored to different CPU architectures, MNN-LLM can better utilize CPU computational power. For instance, the throughput of the \textit{smmla} instruction on ARM i8mm \cite{i8mm} is twice that of \textit{sdot} \cite{i8sdot}. When MNN-LLM detects that the CPU supports i8mm, it rearranges the weights with \( l_p = 8 \) during the model loading phase. This arrangement format enhances performance compared to the data layout in llama.cpp, thereby improving the efficiency of the prefill stage. 

GPUs support hardware loading/storing merging, which allows them to combine a certain number of memory access instructions if the memory addresses accessed by consecutive work items are contiguous. This capability minimizes the number of memory access instructions. Additionally, GPUs can load/store up to 128 bits of data at a time. To maximize memory loading efficiency, each work item should utilize vectorized loading/storing functions. In OpenCL, GPU memory objects are categorized into Buffers and Images. Images can automatically handle boundaries and return appropriate out-of-bounds values based on settings. Certain devices, such as Qualcomm's Adreno GPUs \cite{adreno}, possess powerful texture engines and dedicated L1 caches, enabling efficient loading of data from Image objects. Compared to ordinary buffer objects, Images offer higher bandwidth, making them the preferred choice for storage. To leverage these memory loading advantages, MNN-LLM rearranges GPU weight data and uses Image objects for storage. The rearranged data structure is $[\frac{l}{l_p}, h, l_p]$ with \(l_p = 32\). Each work item loads 4-bit weights at once, totaling 128 bits, which meets the GPU's maximum loading bandwidth and corresponds to the size of four floating-point values in the CL\_RGBA Image memory object. Additionally, each work item accesses data contiguously along the h dimension, ensuring continuous memory reads between work items. Finally, the runtime dynamically adjusts the parallelism based on actual dimensions, allocating a reasonable number of computational tasks to each work item.

For the Attention operator, a similar rearrangement strategy as used for Linear is applied. The key and value are stored directly in the rearranged data layout, ensuring that there is no need to rearrange the historical KV during each computation.

\subsection{Multicore Workload Balancing}

Modern CPUs typically have multiple cores, so effectively utilizing multicore computing capabilities is crucial when optimizing performance. MNN-LLM leverages the multicore parallelism of CPUs to parallelize operations along the $seq len$ and $\frac{h}{h_p}$ dimensions. Considering the big.LITTLE \cite{biglittle} architecture of mobile CPUs, MNN-LLM specifies the computing load for different cores at startup based on their actual computational capabilities. During parallel computation, MNN-LLM allocates the computational workload according to the load rates of the cores. This balancing workload distribution strategy, can enhance multithreaded computing performance compared to the uniform workload strategy.

\begin{figure}[h]
  \centering
  \includegraphics[width=\linewidth]{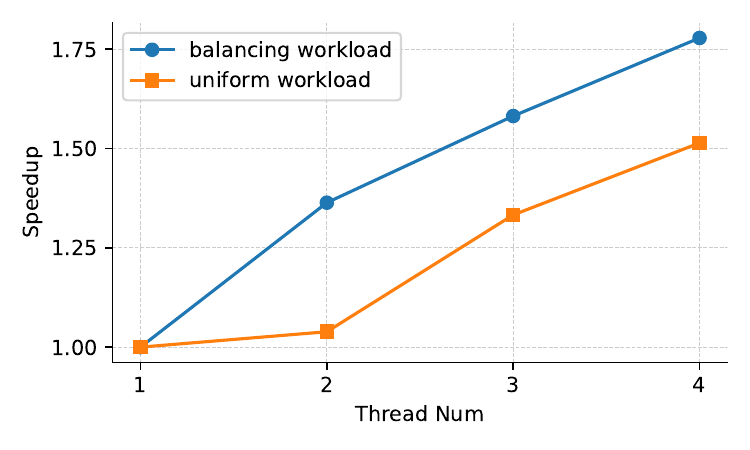}
  \caption{Parallel computing between 1 prime cores and 3 performance cores, speedup achieved through balancing workload and uniform workload.}
  \label{fig:multicore}
\end{figure}

Mainstream mobile SoCs typically feature one prime core and three performance cores, such as the Snapdragon 8 Gen 3 \cite{8gen3}. High-load computations generally utilize the prime core and performance cores. When the number of threads exceeds one, parallel computing between the prime core and performance cores occurs, as shown in Figure~\ref{fig:multicore}. In this scenario, workload balancing significantly improves the multithreaded speedup compared to uniform workload distribution.

\subsection{Mixed Float Precision}

In the previous discussion on matrix operations, low-bit quantization methods were employed to accelerate computations. For non-matrix multiplication operations, MNN-LLM also supports mixed precision for results, ensuring accuracy while enhancing inference performance. ARMv8.2 \cite{armv82} and newer CPUs support float16 calculations, which can save half the memory compared to float32, and the throughput of float16 NOEN \cite{neon} instructions is twice that of float32. However, float16 has some precision limitations; for calculations requiring a higher precision range, significant errors may occur, especially when values exceed 65,504. To address this, MNN-LLM adopts a mixed precision strategy to maintain inference accuracy. During LLM inference, the Softmax calculation in Attention is particularly sensitive to data precision, so MNN-LLM ensures that \textit{Softmax} uses float32. In the matrix multiplication of query and key, the query values may be large, potentially causing overflow after accumulation. To mitigate this, the division by \( \sqrt{d_k} \) \cite{attention} can be applied directly to the query, reducing its value range and preventing overflow in the final result. This approach optimizes overall memory usage and inference performance while maintaining accuracy.

\subsection{Geometry Compute}

The computation graph of LLMs also includes long-tail operators such as \textit{Transpose}, \textit{Gather}, and \textit{Concat}. Although these operators may not significantly contribute to overall execution time, they can result in substantial memory access when data sizes are large. To address these long-tail operators, MNN-LLM employs geometric computation \cite{walle} methods, abstracting all data rearrangement operations as linear mappings of addresses. 
\begin{equation}
  f(\vec{x}) = \vec{offset} + \vec{stride} \vec{x}
\end{equation}
By taking the $\vec{offset}$ and $\vec{stride}$ with a length of 3, we can construct a fundamental mapping relationship described as a Region. This allows us to represent any data rearrangement operators using one or more Regions.

For consecutive data rearrangement operators in the computation graph, this abstraction generates numerous contiguous Regions. MNN-LLM implements an automatic Region Fusion algorithm based on rules like \textit{loop unrolling}, \textit{loop interchange}, \textit{loop tiling}, and \textit{loop fusion}. This algorithm can automatically merge compatible Regions, thereby reducing the number of read and write operations for data rearrangement operators and enhancing performance. By utilizing geometric computation for LLM model inference, the overhead of long-tail operators can be reduced, improving performance by approximately 3\%.

\subsection{LoRA Optimization}

On mobile devices, different tasks may require different LLM models. Due to the large number of model parameters, directly utilizing multiple models can lead to excessive bandwidth and storage usage. Thus, using a base model in conjunction with multiple LoRA models is a more efficient solution for multitasking. 

\begin{table}
  \caption{Computation and Memory under Different LoRA Computation Orders. }
  \label{tab:lora}
  \begin{tabular}{cccc}
    \toprule
    Type & $\left(LoRA_A \cdot LoRA_B\right) \cdot A$ & $LoRA_A \cdot \left(LoRA_B \cdot A\right)$ \\
    \midrule
    Computation & $rh^2 + h^3$ & $2rh^2$ \\
    Memory & $2\left(rh^2 + h^2 + h^3\right)$ & $4rh^2 + hh + rh$ \\
    \bottomrule
  \end{tabular}
\end{table}

MNN-LLM supports the deployment of merged LoRA models and the online loading of multiple LoRA models. When employing multiple LoRA models, MNN-LLM first loads the base model, followed by the computation graph and weights of the LoRA models, with LoRA models sharing the weights of the base model. Given that LoRA weights are generally small, the memory overhead is minimal. Online loading of LoRA models is more flexible than pre-merged approaches, making it suitable for multitasking scenarios, although it incurs additional computational costs. The computation graph for LoRA adds a bypass for the layers involving LoRA, transforming the original computation $A' = W \cdot A$ into $A' = W \cdot A + \left(LoRA_A \cdot LoRA_B\right) \cdot A$, where the additional computations may slow down model inference. Analyzing the characteristics of LoRA weights reveals that the size of R is relatively small compared to the original parameters, allowing us to leverage the associative property of matrix multiplication to alter the computation order, transforming $\left(LoRA_A \cdot LoRA_B\right) \cdot A$ into $LoRA_A \cdot \left(LoRA_B \cdot A\right)$. Assuming the size of matrix A is $\left[h, h\right]$ and that of $LoRA_A$ and $LoRA_B$ is $\left[h, r\right]$, the computation memory access before and after optimization is shown in Table~\ref{tab:lora}. Given that $R$ is relatively small \cite{lora}, rearranging the computation order significantly reduces the memory access volume. For instance, using Qwen2 7B as an example, if $h=3584$ and $r=8$, the optimized memory access volume is only 0.5\% of the original, thereby effectively improving computational efficiency.

\section{Evaluation}
The experimental design hinges upon quantized models, namely Qwen2 1.5B, Qwen2 7B, and Llama3 8B, utilizing the Xiaomi 14 as the test apparatus. Comparative evaluations of inference efficacy are conducted across CPU (harnessing 4 threads) and GPU (via OpenCL) architectures, employing inference engines such as llama.cpp, MLC-LLM, and fastllm. Given that MLC-LLM does not accommodate CPU-based inference and fastllm lacks GPU compatibility, pertinent experiments are excluded for these engines. Extensive trials were executed with prompts of varying lengths (64, 256, and 1024 tokens), with a restrictive upper limit of 16 tokens imposed on the decoding phase.

\begin{figure}[h]
  \centering
  \includegraphics[width=\linewidth]{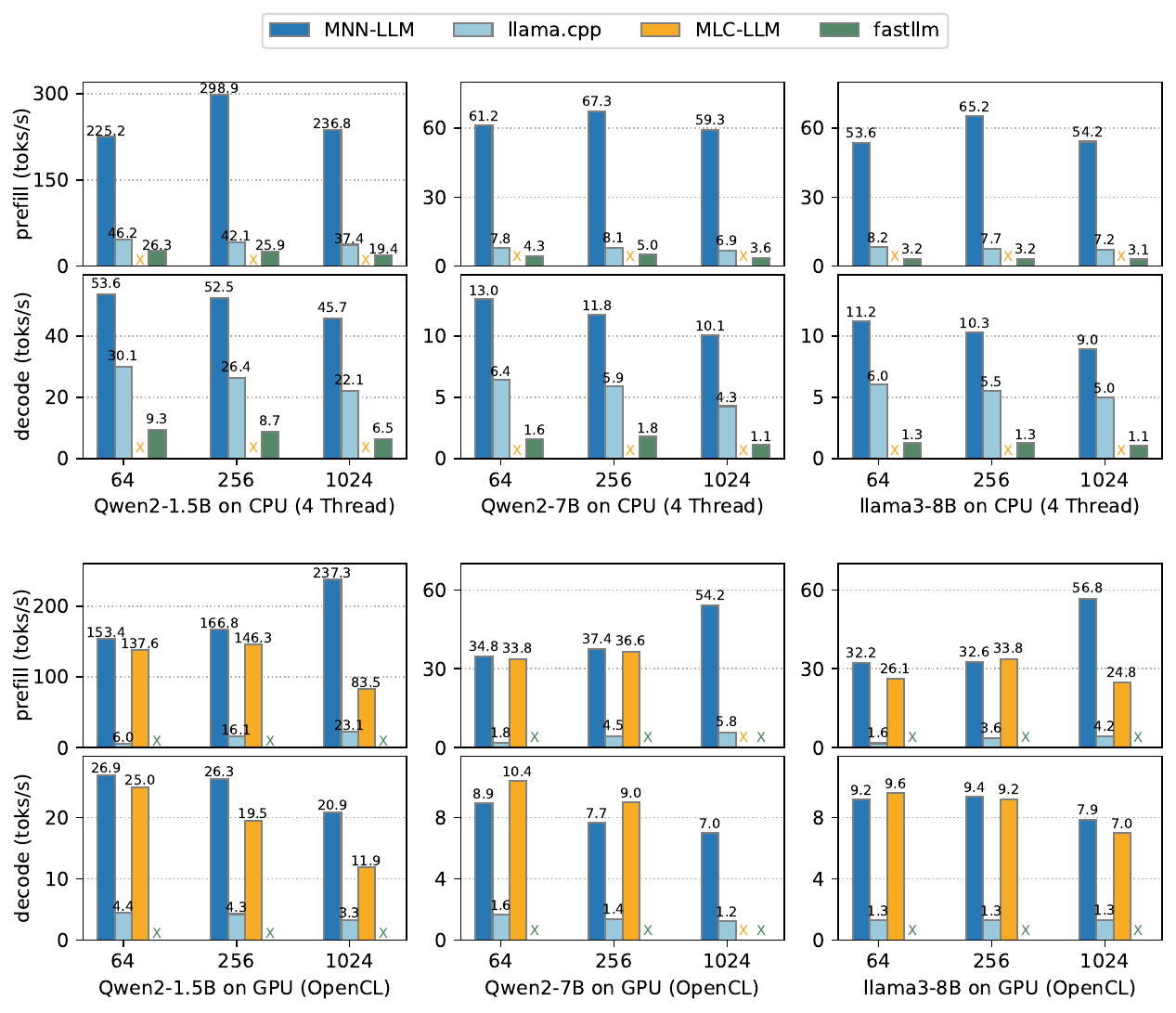}
  \caption{Prefill and decode speeds  of MNN-LLM, llama.cpp, MLC-LLM, and fastllm under different prompt lengths on Xiaomi14's CPUs and GPUs.}
  \label{fig:performance}
\end{figure}

Owing to the poor performances of MLC-LLM in handling asymmetric quantization models, the reported results of MLC-LLM are based on symmetric quantized models but competing engines were explicitly engaged in inference tasks using asymmetric models. The performance results are reflected in terms of the prefill and decode speed, graphically represented in Figure~\ref{fig:performance}.

In CPU benchmarking, MNN-LLM excels, achieving prefill speed boosts of 8.6x over llama.cpp and 20.5x over fastllm, complemented by decoding speeds that are 2.3x and 8.9x faster, respectively. In GPU-based assessments, MNN-LLM's performance slightly declines compared to MLC-LLM, particularly when using Qwen2-7B with shorter prompts, due to MLC-LLM's advantageous symmetric quantization technique. MNN-LLM excels, achieving up to 25.3x faster prefill and 7.1x faster decoding than llama.cpp, and 2.8x and 1.7x improvements over MLC-LLM, respectively.

\section{Conclusion}

This paper introduces MNN-LLM, a high-performance general-purpose inference framework tailored for LLM inference on mobile devices. The framework enhances memory usage through DRAM-Flash Hybrid Storage and Combined Quantization, while improving inference speed with Hardware-Driven Data Reordering, Multicore Workload Balancing, Mixed Float Precision, and Geometry Compute. When compared to leading mainstream frameworks, MNN-LLM achieves up to an 8.6x performance improvement.

% \bibliographystyle{ACM-Reference-Format}
% \bibliography{ref}

\appendix

\end{document}